\documentclass[journal]{IEEEtran}
\usepackage{booktabs}
\usepackage{tabularx}
\usepackage{array}
\usepackage{amsmath}
\usepackage{amssymb}
\usepackage{amsfonts}
\usepackage{graphicx}
\usepackage{epsfig}
\usepackage{subfigure}
\usepackage{psfrag}
\usepackage{cite}
\usepackage{latexsym}
\usepackage{url}
\usepackage{color}
\usepackage{multirow}
\usepackage{mathtools}
\usepackage{bm}
\usepackage{booktabs}
\usepackage{algorithm}
\usepackage{algpseudocode}
\usepackage{bm}
\usepackage[normalem]{ulem} 
\usepackage{verbatim}
\usepackage{indentfirst}
\usepackage{hyperref}
\usepackage{multirow}
\usepackage{bbm}
\usepackage{booktabs}
\usepackage{threeparttable}

\graphicspath{{fig/}}

\PassOptionsToPackage{bookmarks={false}}{hyperref}

\IEEEoverridecommandlockouts

\allowdisplaybreaks[4] 

\begin{document}
\title{Wireless AI Networks: Co-Adapting Communication Resources and Distributed Intelligence}
\author{Ziqiang Ye,\IEEEmembership{~Student Member, IEEE,~}
Yue Xiao,\IEEEmembership{~Member, IEEE,~}
Qiyan Zhan,\IEEEmembership{~}\\
Yulan Gao,\IEEEmembership{~Member, IEEE,~}
Xianfu~Lei,~\IEEEmembership{Member,~IEEE,~}
and Vladimir Poulkov,\IEEEmembership{~Senior Member, IEEE}
\thanks{Z. Ye, Q. Zhan and Y. Xiao are with the National Key Laboratory of Wireless Communications, University of Electronic Science and Technology of China, Chengdu 611731, China (email: yysxiaoyu@hotmail.com, qyzhan@std.uestc.edu.cn, xiaoyue@uestc.edu.cn).}
\thanks{Y. Gao is with the Division of Information Science and Engineering, KTH Royal Institute of Technology, Sweden (email: yulang@kth.se). }
\thanks{X. Lei is with the School of Information Science and Technology, Institute of Mobile Communications, Southwest Jiaotong University, Chengdu 610031, China (e-mail: xflei@home.swjtu.edu.cn).}
\thanks{V. Poulkov is with the Faculty of Telecommunications, Technical University of Sofia, 1000 Sofia, Bulgaria (e-mail: vkp@tu-sofia.bg).}
}

\maketitle

\begin{abstract}
Distributed artificial intelligence (AI) increasingly relies on wireless networks, where communication and computing resources shape how AI tasks are configured and executed. However, resource availability alone does not determine how effectively these tasks can be supported, since different information selections and transmission strategies may yield different AI performance under the same resource constraints. To address this challenge, this article presents Wireless AI Networks as a unified design perspective for bidirectional adaptation between wireless communication and distributed intelligence. Specifically, an AI-executable resource budget characterizes candidate AI configurations, while task-level information value guides their selection and wireless resource allocation. Building on these principles, a wireless-AI co-adaptation framework coordinates resource abstraction, joint configuration and allocation, and execution feedback. Two complementary experiments illustrate this perspective: update selection and statistical correction improve federated learning under identical model-update payload budgets, while sensitivity-aware power allocation reduces model accuracy degradation under the same average transmit-power constraint. Together, these results highlight the importance of coordinating resource feasibility and task-level information value in distributed AI execution.
\end{abstract}

\section{Introduction}

Artificial intelligence (AI) is increasingly deployed across mobile devices, edge servers, connected vehicles, robots, and other networked agents \cite{liu2026edge}.
As a result, AI training and inference increasingly depend on information exchange among distributed nodes. Federated learning, collaborative inference, foundation-model services, and multi-agent systems rely on wireless connectivity to exchange model updates, intermediate features, and task-related information \cite{jiang2026large}.
In these settings, variations in wireless connectivity and computing resources influence both information delivery and AI execution.
For example, limited uplink capacity may restrict the model updates available for federated aggregation, while communication latency and edge computing capacity may determine the feasible partition points of an inference model.
Consequently, supporting distributed intelligence requires coordinated adaptation of wireless communication and AI execution to changing resource conditions.

To accommodate these resource variations, communication and computing constraints have been incorporated into distributed AI design. In particular, communication-aware federated learning adapts client participation and model-update transmission to available network resources \cite{wang2019adaptive,yang2019scheduling}, while collaborative inference adjusts model partitioning and intermediate representations to communication and computing conditions \cite{kang2017neurosurgeon}.
These approaches demonstrate the benefits of adapting AI execution to resource availability. However, their resource representations, adaptation decisions, and performance criteria are typically tailored to individual applications. A unified design perspective can therefore help organize these task-specific approaches by connecting network resource constraints with AI execution requirements. Such a perspective should account for both which AI configurations are supportable and how the information exchanged within those configurations contributes to task performance.

Motivated by these considerations, this article presents Wireless AI Networks as a unified design perspective built on resource feasibility and task-level information value. The former characterizes which AI operations can be supported by the available communication and computing resources. To express these constraints in terms meaningful to AI execution, an AI-executable resource budget is introduced, specifying task-relevant operating limits such as admissible update sizes, model partitions, and device participation. However, multiple configurations may satisfy these limits while yielding different AI performance. Task-level information value therefore provides a complementary basis for selecting among candidate configurations and allocating transmission resources. Depending on the task, this value may be estimated from model-update characteristics, intermediate-feature relevance, or the sensitivity of model accuracy to transmission errors.

These two considerations give rise to complementary directions of wireless-AI adaptation. In the wireless-to-AI direction, AI configurations, including update selection, model partitioning, and device participation, are adjusted to available network resources. Conversely, in the AI-to-wireless direction, the importance and error sensitivity of AI information guide transmission decisions, such as power allocation, scheduling, and reliability protection. These decisions are interdependent, since the selected AI configuration determines transmission requirements, while wireless allocation affects configuration feasibility and execution performance. Their coordination therefore requires joint consideration of resource constraints and task performance. Furthermore, feedback from communication and AI execution informs subsequent decisions as network conditions and task requirements evolve.

The main contributions of this article are summarized as follows:

\begin{itemize}
    \item Wireless AI Networks is presented as a unified design perspective connecting resource feasibility with task-level information value. Specifically, an AI-executable resource budget maps communication and computing constraints to candidate AI configurations, while task-dependent value estimates guide information selection and transmission.

    \item Building on this perspective, a wireless-AI co-adaptation framework is developed to coordinate AI configuration and wireless resource allocation under common resource and performance constraints. The framework integrates resource abstraction, joint decision making, and execution feedback, thereby supporting adaptation to changing network conditions and task requirements.

    \item Finally, two complementary experiments illustrate the adaptation principles through partial model updates in wireless federated learning and sensitivity-aware model transmission. The results demonstrate that update selection and statistical correction improve learning performance under identical model-update payload budgets, while sensitivity-aware power allocation reduces accuracy degradation under the same average transmit-power constraint.
\end{itemize}

The remainder of this article is organized as follows. Section II introduces the Wireless AI Networks design perspective and its resource abstraction. Building on these concepts, Section III presents the wireless-AI co-adaptation framework, while Section IV illustrates its design principles through two complementary experiments
Finally, Section V concludes the article and outlines future research directions.

\section{From AI over Wireless to Wireless AI Networks}

Wireless connectivity enables distributed AI nodes to exchange information and coordinate computation across devices, edge servers, and cloud platforms \cite{luo2026ai}.
However, changes in communication and computing resources affect which AI operations can be supported and how effectively they can be executed.
This section examines these relationships and introduces resource feasibility and task-level information value as complementary considerations in wireless-AI design.

\subsection{From Communication Efficiency to Task-Oriented Adaptation}

\begin{figure*}[!t]
  \centering
  \includegraphics[width=\textwidth]{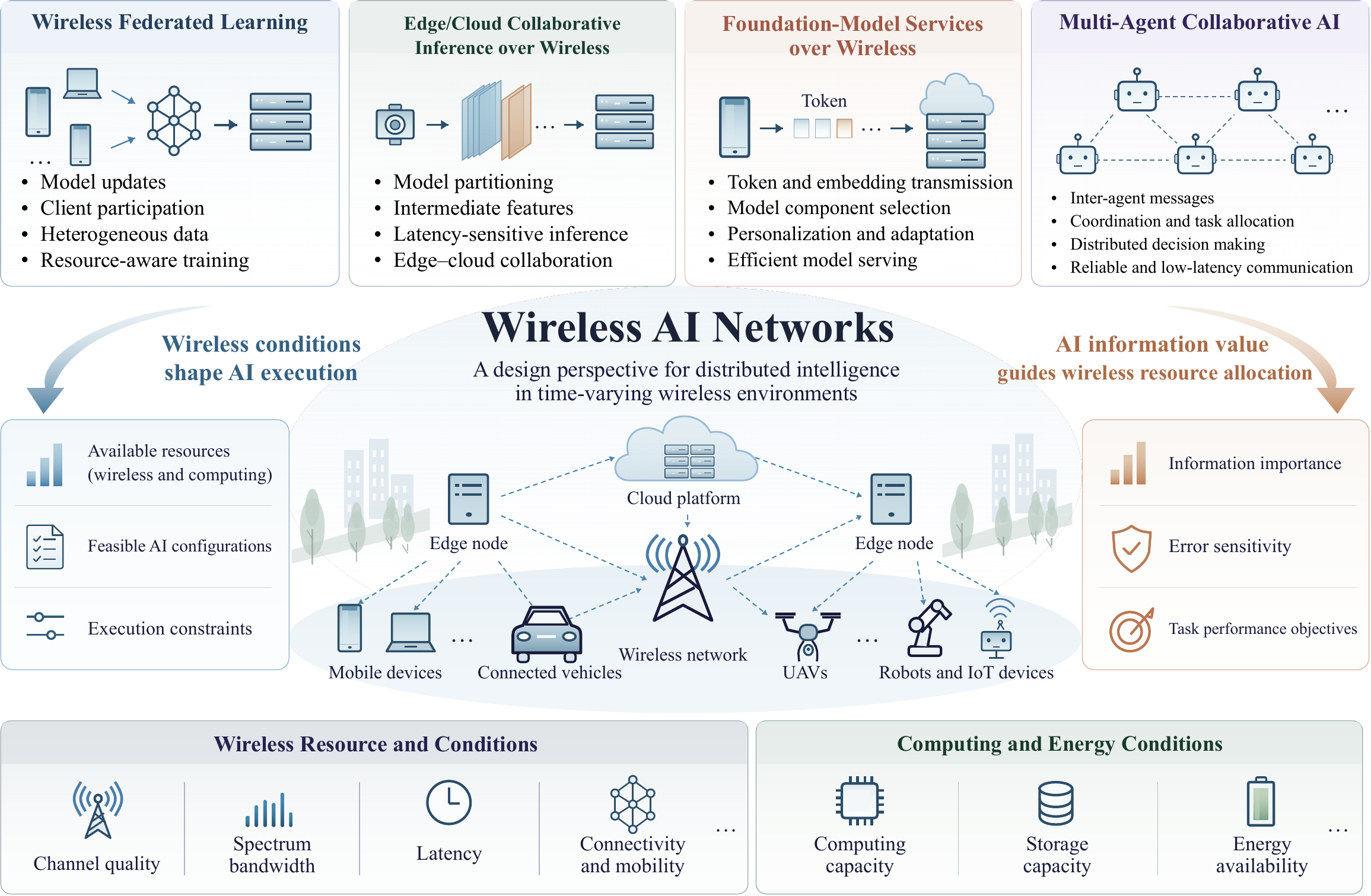}
  \caption{Overview of Wireless AI Networks as a design perspective for distributed intelligence. Wireless conditions shape feasible AI execution, while task-level information value guides wireless resource allocation across federated learning, collaborative inference, foundation-model services, and multi-agent systems.}
  \label{fig:1}
\end{figure*}

Communication-efficient AI over wireless seeks to reduce the cost of exchanging information through techniques such as model compression, sparsification, and quantization \cite{wang2025empowering,lin2017deep,alistarh2017qsgd}.
Beyond reducing payload size, communication-aware federated learning adapts client participation and model-update transmission to network conditions \cite{paolini2026programmable,amiri2020federated}, while collaborative inference adjusts model partitioning and intermediate representations to available communication and computing resources \cite{hao2026dnn,shao2020bottlenet++}.
Although these approaches address different tasks, they share a common requirement: translating network resource constraints into AI execution decisions.

This translation connects two different descriptions of the same system. Wireless networks are characterized by channel quality, available bandwidth, achievable rate, and connectivity, together with latency, reliability, and energy constraints.
AI execution, in turn, is configured through model components, feature representations, participating devices, and computation placement.
Consequently, a resource description becomes useful to an AI task when it identifies the operations and information exchanges that can be supported.
For example, an uplink transmission budget may be translated into an admissible model-update size, while communication and computing constraints jointly restrict the candidate partition points of an inference model.

Resource feasibility alone, however, does not determine which configuration should be selected.
Different update subsets or feature representations may satisfy the same resource limits yet contribute differently to learning or inference performance. Moreover, their sensitivity to transmission errors may differ, leading to different requirements for wireless protection. Task-oriented adaptation therefore requires both identifying supportable AI configurations and evaluating the task-level consequences of selecting and transmitting their information. These complementary considerations motivate the resource abstraction and information-value perspective introduced next.

\subsection{Wireless AI Networks and Task-Aware Resource Abstraction}

Wireless AI Networks describes a design perspective in which communication and computing conditions shape AI execution, while AI task requirements and information characteristics guide wireless resource allocation. As illustrated in Fig.~\ref{fig:1}, this perspective applies to distributed learning, collaborative inference, foundation-model services, and multi-agent systems. Across these applications, the common objective is to coordinate the information required by an AI task with the resources available for its processing and delivery.
To establish this connection, an AI-executable resource budget is introduced to express network resource constraints as task-relevant operating limits. The abstraction accounts for payload and signaling overhead, transmission deadlines, reliability requirements, and processing demands. Accordingly, available resources can be mapped to candidate AI configurations, including update sizes and client participation in federated learning, or model partitions and feature dimensions in collaborative inference. Since transmission requirements depend on the selected configuration, these candidates remain subject to the wireless allocation and transmission mode determined during joint decision making.
Consider, for example, a federated-learning client whose effective uplink budget permits the transmission of only 2\% of its model coordinates. After accounting for representation and signaling overhead, multiple coordinate subsets may satisfy this limit. Nevertheless, random selection and update-dependent sampling can produce different aggregation outcomes despite transmitting the same number of coordinates. The budget therefore characterizes the admissible update size, while the selection and aggregation strategy determines how effectively that budget supports learning.

Task-level information value provides a basis for distinguishing among these alternatives \cite{shao2021learning}.
Its estimation depends on both the AI task and the decision under consideration. Model-update characteristics may guide coordinate selection, whereas validation-based sensitivity estimates may indicate how corruption of different model bits affects accuracy. Such estimates represent task-dependent consequences rather than a universal information-value metric. They can therefore guide both the selection of AI information and the allocation of transmission resources, including power and reliability protection \cite{jankowski2024airnet}.
Together, resource feasibility and task-level information value form the basis of the wireless-AI co-adaptation framework developed in the following section.

\section{Wireless-AI Co-Adaptation Framework}
\label{sec:framework}

\begin{figure*}[!t]
  \centering
  \includegraphics[width=\textwidth]{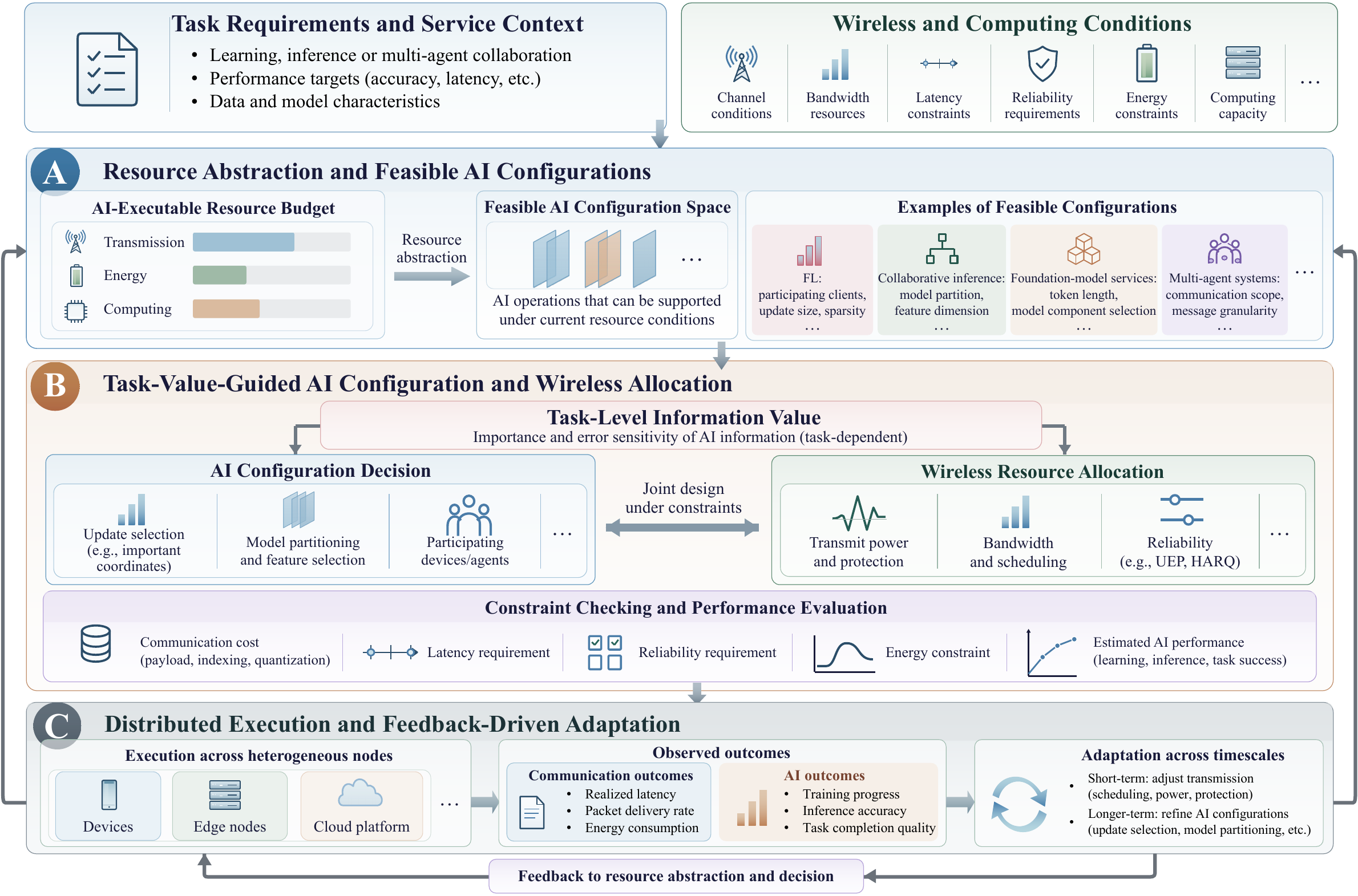}
  \caption{Wireless-AI co-adaptation framework. Stage A derives candidate AI configurations from the resource envelope. Stage B jointly refines AI configuration and wireless allocation, revisiting Stage A when feasibility constraints require an updated candidate set. Stage C uses observed communication and AI outcomes to support subsequent adaptation.}
  \label{fig:2}
\end{figure*}

Building on the resource abstraction and information-value perspective introduced in Section II, Fig.~\ref{fig:2} presents a wireless-AI co-adaptation framework comprising three stages: resource abstraction and candidate AI configuration identification, joint AI configuration and wireless resource allocation, and distributed execution with feedback. Together, these stages connect current resource conditions with task-dependent decisions and observed execution outcomes. The interaction between Stages A and B refines candidate configurations before execution, while Stage C provides feedback for subsequent adaptation.

\subsection{Resource Abstraction and Candidate AI Configurations}

The first stage establishes a resource envelope from current wireless and computing conditions, together with task requirements. Channel estimates, available bandwidth and power, energy budgets, and computing capacity characterize the available resources, while transmission deadlines and reliability requirements constrain their use. Candidate transmission modes are then considered to translate this envelope into an AI-executable resource budget, accounting for payload representation, signaling overhead, and processing demands.

The resulting budget is expressed through task-specific configuration variables. In federated learning, for example, it restricts admissible combinations of client participation and update size. In collaborative inference, it identifies candidate model partitions and intermediate-feature dimensions that can be supported by communication and computing resources. The output is therefore an initial set of candidate AI configurations whose final feasibility depends on the wireless allocation and transmission protection determined in Stage B. If joint evaluation reveals that a candidate cannot meet the resource or performance constraints, the operating limits are revised and the candidate set is updated accordingly.

\subsection{Task-Value-Guided AI Configuration and Wireless Allocation}

Given the candidate configurations identified in Stage A, the second stage jointly refines AI configuration and wireless resource allocation according to task-level information value. Specifically, the AI configuration determines which information is processed or exchanged, while wireless allocation determines how that information is delivered. Since these decisions jointly affect resource consumption and task performance, they are evaluated under common communication, computing, and performance constraints.

Information-value estimates are supplied by the AI task, whereas channel and resource estimates are provided by wireless resource management. In federated learning, for example, client-side update statistics can guide coordinate selection, together with the statistical correction required for aggregation. For quantized model transmission, validation-based sensitivity estimates can guide power allocation or reliability protection across bit positions. An edge-side task controller can coordinate these inputs with the wireless scheduler, with the resulting decisions implemented at the relevant devices and radio nodes.

Each candidate configuration and its associated resource assignment are evaluated in terms of communication cost, processing demand, latency, reliability, energy consumption, and estimated AI performance. Communication cost includes the transmitted values and any required indexing, quantization metadata, and control signaling. Consequently, a configuration that satisfies a nominal payload limit may still be infeasible under a particular transmission or protection strategy.

If the requirements cannot be met within the current candidate space, Stage B returns updated operating restrictions to Stage A so that the candidate set can be revised. Otherwise, the selected AI configuration and wireless allocation proceed to execution. This interaction refines feasibility within the current decision cycle, while the execution feedback introduced next informs subsequent adaptation.

\subsection{Distributed Execution and Feedback-Driven Adaptation}

The third stage executes the selected AI configurations and transmission strategies across devices, edge nodes, and cloud platforms.
Observed transmission latency, packet delivery, and energy consumption are collected alongside training progress, inference accuracy, or task-completion quality.
These observations update both resource estimates and subsequent decisions.

Feedback operates across different timescales.
Channel and delivery measurements update the resource envelope and transmission estimates, supporting short-term adjustments to scheduling, power, and protection.
AI-side outcomes refine task-dependent value estimates and configuration policies, including update-selection probabilities, model partitioning, and bit-sensitivity profiles.
Execution feedback therefore revises the inputs to Stages A and B for subsequent decisions, while the B-to-A feasibility revision operates within the current decision cycle.

\section{Case Study}

Two complementary experiments illustrate wireless-AI co-adaptation through model-information exchange over wireless links. The first examines how time-varying uplink resources constrain partial model updates in federated learning (FL), whereas the second investigates how the error sensitivity of quantized model information guides transmission power allocation. Together, they illustrate resource-driven AI configuration and task-guided wireless allocation.

\subsection{Wireless AI Scenario and Experimental Setup}

For wireless-to-AI adaptation, ten clients collaboratively train a lightweight convolutional neural network on CIFAR-10. Local datasets follow a non-IID Dirichlet partition with $\alpha=0.3$.
Training proceeds for 100 communication rounds, with one local epoch per admitted client. In each round, seven clients are selected uniformly without replacement, giving each client an admission probability of $0.7$.
The remaining clients do not upload updates in that round.

To represent time-varying wireless opportunities, each client’s uplink SNR is independently sampled in each round from a Gaussian distribution in decibels with mean $12$~dB and standard deviation $4$~dB, truncated to $[3,21]$~dB.
Each admitted client is allocated orthogonal uplink resources with a bandwidth of $180$~kHz and a $200$~ms transmission window.
The model-update payload budget is obtained from the instantaneous Shannon rate using an effective-rate factor of $0.8$.
This budget counts FP32 update values and excludes coordinate indices and control signaling. At the nominal operating point, the average uploaded coordinate ratio is $2.336\%$; the budget sweep varies this average while retaining the same SNR model. Thus, client admission and channel-dependent payload budgets jointly constrain the information available for aggregation.

Channel-adaptive unbiased partial-update federated learning (CAUP-FL) is adopted to accommodate these constraints. For each admitted client, the payload budget determines the number of coordinates to upload. Coordinate inclusion probabilities are then constructed from update magnitudes subject to a minimum coverage constraint, and a fixed-cardinality subset is sampled accordingly. Before transmission, the selected values are corrected by inverse-probability weighting at the client, so the inclusion probabilities need not be communicated. The server additionally accounts for client admission probabilities during aggregation. In this way, channel-adaptive update selection is combined with statistical correction.

CAUP-FL is compared with five partial-update baselines. Dynamic Random selects coordinates uniformly, with the upload size determined by the instantaneous payload budget. Fixed Random and Fixed Top-k impose predefined upper limits on update size and reduce the number of uploaded coordinates when the instantaneous budget is insufficient. By contrast, Static-rate Random and Static-rate Top-k use upload sizes determined by long-term channel statistics and skip transmission whenever the instantaneous budget cannot accommodate the prescribed update. Random-based schemes select coordinates uniformly, whereas Top-k schemes retain the largest-magnitude update coordinates. The two static-rate schemes are included in the learning-curve comparison. Dense FedAvg transmits complete model updates and serves as a communication-intensive reference.

For the complementary AI-to-wireless direction, $8$-bit uniformly quantized CNN parameters are transmitted over independent Rayleigh flat-fading channel uses with coherent binary phase-shift keying (BPSK) and perfect receiver channel knowledge. Quantized weights use a signed two’s-complement representation. The power-allocation strategies are evaluated at SNRs from $6$ to $12$~dB under a common average transmit-power constraint, using $200$ independent channel realizations per SNR point.

\subsection{Wireless-to-AI Adaptation: Resource-Aware Model Updates}

\begin{figure*}[!t]
  \centering
  \includegraphics[width=\textwidth]{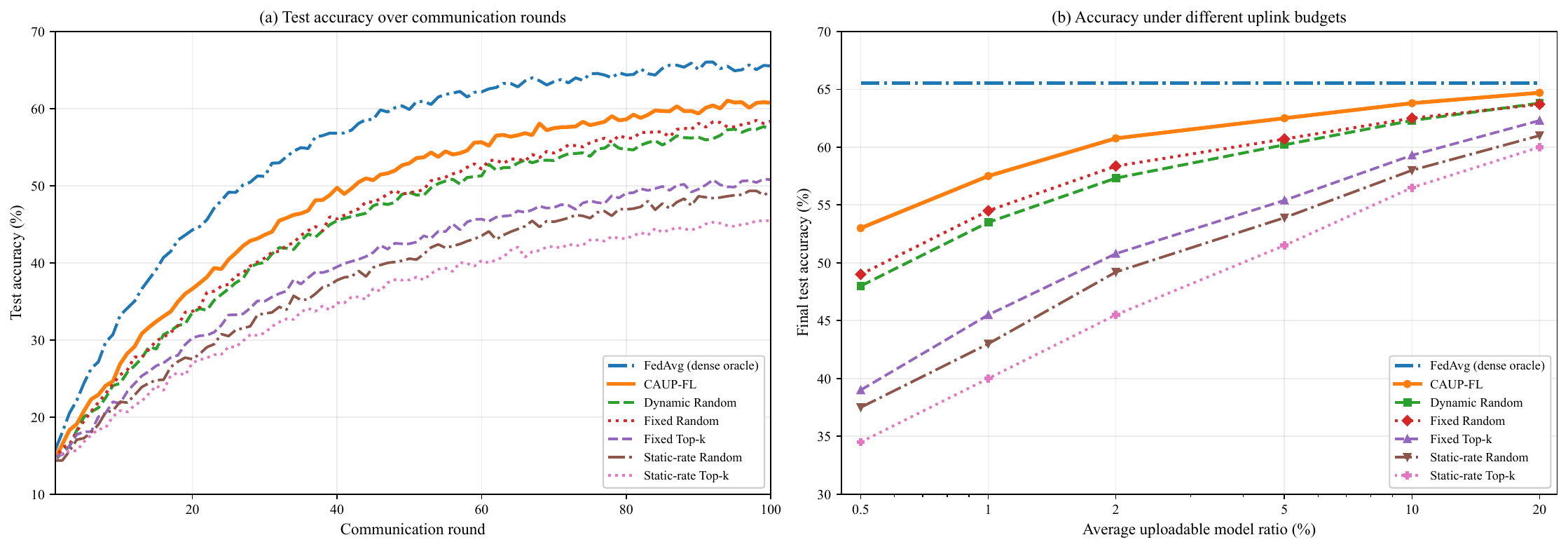}
  \caption{Wireless-to-AI adaptation through resource-aware model updates: (a) test accuracy over communication rounds and (b) final test accuracy under different uplink budgets.}
  \label{fig:3}
\end{figure*}

\begin{table}[t]
\centering
\caption{CIFAR-10 results over 12 paired seeds, Dirichlet $\alpha=0.3$, and 30\% admission outage. Accuracy is reported as mean $\pm$ standard deviation.}
\label{tab:caupfl_measured_case}
\resizebox{\columnwidth}{!}{%
\begin{threeparttable}
\begin{tabular}{lccc}
\toprule
Method & Accuracy (\%) & Payload (Mbit) & Coordinates (\%) \\
\midrule
FedAvg (dense oracle) & $65.54\pm2.42$ & 3025.344 & 100.000 \\
Fixed Top-$k$ & $50.79\pm3.22$ & 70.664 & 2.336 \\
Fixed Random & $58.36\pm4.77$ & 70.664 & 2.336 \\
Dynamic Random & $57.32\pm5.70$ & 70.664 & 2.336 \\
\textbf{CAUP-FL} & $\mathbf{60.76\pm4.58}$ & 70.664 & 2.336 \\
\bottomrule
\end{tabular}
\begin{tablenotes}[flushleft]\footnotesize
\item Payload counts FP32 model-update values and excludes coordinate indices and control signaling. CAUP-FL and Dynamic Random have identical payloads within each paired seed. Their paired accuracy difference is $+3.43$ percentage points (95\% interval $[+1.61,+5.26]$).
\end{tablenotes}
\end{threeparttable}%
}
\end{table}

Fig.~\ref{fig:3}(a) compares the learning trajectories over 100 communication rounds, while Table~\ref{tab:caupfl_measured_case} summarizes final test accuracy across 12 paired seeds.
CAUP-FL achieves $60.76\%$ mean accuracy, compared with $57.32\%$ for Dynamic Random.
The paired improvement is $3.43$ percentage points, with a 95\% interval of $[+1.61,+5.26]$ percentage points.

At the nominal operating point, CAUP-FL and Dynamic Random have the same reported model-update payload of $70.664$~Mbit.
Their performance difference reflects the combined effects of update-dependent coordinate selection and statistical correction rather than additional uploaded model values.
Dense FedAvg achieves $65.54\%$ accuracy with a substantially larger payload of $3025.344$~Mbit.

Fixed Random achieves $58.36\%$ accuracy, compared with $57.32\%$ for Dynamic Random, although both employ uniform coordinate selection and adjust upload sizes to instantaneous channel conditions.
CAUP-FL achieves higher accuracy by combining channel-adaptive upload sizes with update-dependent coordinate selection and statistical correction.
Fig.~\ref{fig:3}(b) further shows that its advantage persists across the evaluated upload ratios, with larger gains under tighter budgets.
These results illustrate how the selection and delivery of model information affect learning performance under constrained wireless resources.

\subsection{AI-to-Wireless Adaptation: Sensitivity-Aware Transmission}
\label{subsec:ai_to_wireless}

For quantized model parameters, bit errors at different positions have unequal effects on the reconstructed model and its task performance.
We estimate the sensitivity of each of the eight bit planes by perturbing the corresponding bits in a copy of the quantized model and measuring the resulting validation-accuracy loss.

For each bit plane, we flip that bit in $1\%$ of randomly selected quantized weights and average the validation-accuracy loss over ten random masks.
Transmit power is allocated in proportion to the measured sensitivity, with a positive offset of $10^{-4}$, and normalized to the same average bit power across all schemes.
For comparison, Equal assigns uniform power to all bit planes, while UEP-3dB and UEP-6dB provide bit planes $4$--$7$ with fixed $3$ and $6$~dB power advantages over planes $0$--$3$, respectively.
All schemes satisfy the same average power constraint.

\begin{figure}[!t]
  \centering
  \includegraphics[width=\columnwidth]{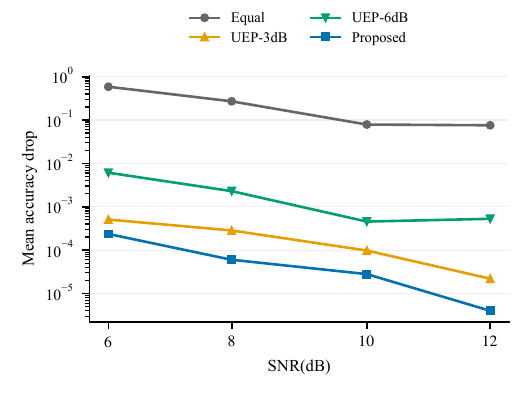}
  \caption{AI-to-wireless adaptation through sensitivity-aware bit-level power allocation. Mean accuracy drop versus signal-to-noise ratio (SNR) under equal power, fixed unequal error protection (UEP), and sensitivity-aware allocation, with the same average transmit-power budget.}
  \label{fig:4}
\end{figure}

The accuracy drop is defined as the nonnegative difference between the test accuracy of the error-free quantized model and that of the received model, expressed as a fraction.
Fig.~\ref{fig:4} reports the mean accuracy drop over $200$ independent channel realizations at each SNR.
Equal power allocation yields the largest accuracy loss, while both fixed UEP schemes reduce the degradation.
UEP-3dB outperforms UEP-6dB, indicating that a larger fixed power imbalance does not necessarily improve model protection.
Sensitivity-aware allocation achieves the lowest mean accuracy drop across the evaluated SNR range by assigning greater protection to bit planes with higher measured impact on model performance.

The two experiments illustrate complementary adaptation decisions: wireless resource availability determines the model information that can be uploaded for learning, while task-dependent error sensitivity guides the protection of transmitted AI information.

\section{Conclusion and Future Directions}
\label{sec:conclusion}

This article has presented Wireless AI Networks as a unified design perspective connecting wireless resource availability with distributed AI execution. An AI-executable resource budget identifies candidate AI configurations, while task-level information value guides their joint refinement with wireless resource allocation. Two complementary experiments have illustrated these principles. Specifically, update selection and statistical correction improve federated learning under identical model-update payload budgets, while sensitivity-aware power allocation reduces model accuracy degradation under a common average transmit-power constraint. These findings motivate the following research directions.

\textbf{1) Task-Level Information Value Estimation and Resource Abstraction:}
The contribution of exchanged information depends on both the AI task and its current execution state. Efficient estimation methods are therefore needed to relate update characteristics, feature relevance, and error sensitivity to task performance. Moreover, the computation and signaling costs of obtaining these estimates should be incorporated into the resource abstraction, so that adaptation gains can be assessed against their implementation overhead.

\textbf{2) Multi-Timescale Adaptation with Reconfiguration Overhead:}
Wireless conditions may vary faster than AI configurations and information-value estimates. Consequently, rapid adjustments to scheduling, power, and protection must be coordinated with slower changes in model partitioning, participation policies, and configuration strategies. Future designs should determine when reconfiguration is beneficial while accounting for signaling cost, switching overhead, and decision stability.

\textbf{3) Cross-Task Value Comparison and Network-Wide Coordination:}
Concurrent AI services may quantify information value through different measures, including learning progress, inference accuracy, and task success. Allocating shared bandwidth, power, and computing resources therefore requires comparable task-level utility signals that also reflect service priorities and performance requirements. Developing such mappings would extend the proposed perspective from individual tasks to the coordination of heterogeneous AI workloads.

\bibliographystyle{ieeetr}
\bibliography{references}

\end{document}